\documentclass[preprint,review,12pt]{elsarticle}
\usepackage{xcolor}

\usepackage{comment}
\usepackage[utf8]{inputenc}
\usepackage{amsmath,amssymb,amsfonts}
\usepackage{graphicx}
\usepackage{url} 
\usepackage{subfig}
\usepackage[ruled,vlined]{algorithm2e}
\usepackage{multirow}

\journal{Computers \& Industrial Engineering}

\begin{document}

\begin{frontmatter}



\title{Coverage Path Planning for Spraying Drones}


\author[inst1]{E. Viridiana Vazquez-Carmona}

\affiliation[inst1]{organization={Instituto Politécnico Nacional (IPN), Centro de Innovación y Desarrollo Tecnológico en Cómputo (CIDETEC), },                 addressline={Av. Luis Enrique Erro S/N},
            city={Ciudad de México},
            postcode={07738},
            country={México}}
\author[inst1]{Juan Irving Vasquez \corref{cor1}}

\cortext[cor1]{Corresponding author \\ }
\ead{jvasquezg@ipn.mx}

\author[inst1]{Juan Carlos Herrera-Lozada}


\author[inst3]{Mayra Antonio-Cruz}
\affiliation[inst3]{organization={Instituto Politécnico Nacional, UPIICSA, SEPI},
            addressline={Av. Té 950, Granjas México, Iztacalco}, 
            city={Mexico City},
            postcode={08400}, country={Mexico}}

\begin{abstract}
The pandemic by COVID-19 is causing a devastating effect on the health of the global population. Currently, there are several efforts to prevent the spread of the virus. Among those efforts, cleaning and disinfecting public areas have become important tasks and they should be automated in future smart cities. To contribute in this direction, this paper proposes a coverage path planning algorithm for a spraying drone, an unmanned aerial vehicle that has mounted a sprayer/sprinkler system, to disinfect areas. \textcolor{black}{State-of-the-art planners consider a camera instead of a sprinkler, in consequence, the expected coverage will differ at running time because the liquid dispersion is different from a camera's projection model. In addition, current planners assume that the vehicles can fly outside the target region; this assumption can not be satisfied in our problem, because disinfections are performed at low altitudes. Our method presents i) a new sprayer/sprinkler model that fits a more realistic coverage volume to the drop dispersion and ii) a planning algorithm that efficiently restricts the flight to the region of interest avoiding potential collisions in bounded scenes. The algorithm has been tested in several simulation scenes, showing that \textcolor{black}{it} is effective and covers more areas with respect to other two approaches in the literature.} Note that the proposal is not limited to disinfection applications, but can be applied to other ones, such as painting or precision agriculture.
\end{abstract}



\end{frontmatter}

\section{Introduction}

\begin{figure}[tb]
		\centering
		\includegraphics[width=\linewidth]{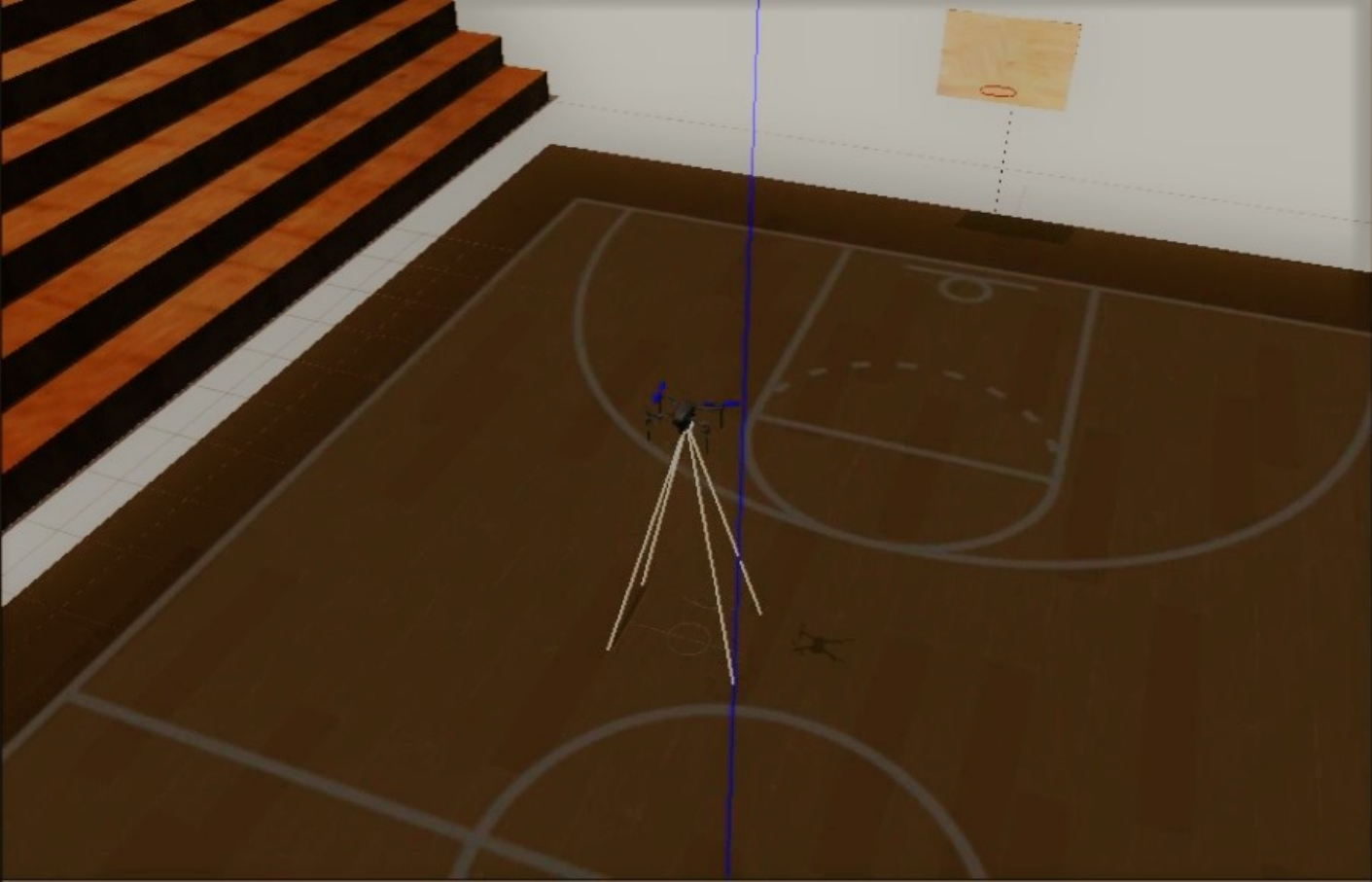}
		\caption{Software in the loop simulation of a basketball court disinfection using the proposed method and Gazebo simulator.}
		\label{fig:simulation}
\end{figure}

The pandemic by COVID-19 is causing a devastating effect on global public health. Because of that, various efforts have been made to face it. Among those efforts, cleaning and disinfection of public areas have been important. Therefore, tools that facilitate the task are needed. For this purpose, in the automation field, some tools have been provided, example of that are the diagnostic systems using computer vision and artificial intelligence techniques, human support robots for the cleaning and maintenance of door handles, and robotic automation for distribution of food and essentials \cite{Bhargava,Ramalingam,Sharma}. Another example is the disinfection task, which involves spreading a liquid that cleans surfaces. So far, in the majority of cases, it is performed by human workers that put their health at risk. To carry out the disinfection task in an autonomous fashion, a positioning system or robot moves an active sprinkler in such a way all \textcolor{black}{Regions of Interest} (ROI) are covered \cite{jaber}. Within positioning systems, \textcolor{black}{Unmanned Aerial Vehicles (UAVs)}, also known as drones, can move in three dimensions while they have a relatively low cost. In consequence, relatively large areas such as buildings, courts or halls can be disinfected quickly. See Fig. \ref{fig:simulation} as an example of the disinfection of a basketball court.

Autonomous disinfection requires the integration of several techniques into one single application that is executed before the flight and also during it. First, the sprinkler coverage \textcolor{black}{needs to be modeled.} Next, the ROI \textcolor{black}{has to be} specified by defining a polygon in geodetic coordinates. Then the \textcolor{black}{unmanned aerial vehicle (UAV)} is prepared for the mission. Next, the flight mission is planned. Finally, the UAV automatically \textcolor{black}{has to take off and follows} the planned path. Among these tasks, this paper focuses on the planning technique, which is carried out before flying and receives as input the ROI, the sprinkler system and the vehicle's capabilities in order to determine a set of waypoints that will be loaded in the on-board controller, so that in a next stage, the vehicle \textcolor{black}{flies} autonomously following the planned path. The core problem in the planning stage is the \textcolor{black}{Coverage Path Planning} (CPP)\textcolor{black}{; which} is defined as determining a path for a vehicle so that the ROI is covered \textcolor{black}{\cite{galceran}}. In addition, planning for disinfection requires adequate modeling of the sprinkler system, otherwise, disinfection is compromised.


After reviewing the literature on CPP for UAVs and spraying UAVs \textcolor{black}{(whose detailed analysis is presented in section \ref{sec:related work})}, two fashions were found: i) The one focusing on solving the CPP problem for UAVs, and ii) the second one concerning the spraying task without considering the CPP problem. However, the first fashion is mainly dedicated to applications on surveillance, obstacle avoidance, inspection, mapping, and reconnaissance of areas. Only four papers were dedicated to the CPP problem addressing the spraying task \cite{me2016quadcopter,GAO2020,Yao2016,Muliawan2019}. In general, for crop areas, they do not consider the model of the sprinkler \textbf{neither} the collisions that \textbf{may} occur in the surrounding area. Also, those papers do not model the footprint of the sprinkler and some are limited to a certain shape of the terrain to perform the spraying task. In order to contribute in this direction, particularly to provide a spraying system for the autonomous disinfection useful to face the pandemic by COVID-19, this paper integrates the aforementioned fashions, developing a CPP method that allows disinfecting convex regions with \textcolor{black}{a} UAV. 

\textcolor{black}{
The proposed method has three advantages with respect to current methods: i) it integrates a sprinkler model, ii) the planned paths do not go outside the target region, and iii) the computation time is shorter with respect to the compared methods. To achieve those advantages, the method makes improvements over the algorithms introduced in \cite{torres2016coverage,vasquez2018coverage,vasquez2020coverage}. First, the sprinkler is modeled as a 3D paraboloid whose intersection with the ground creates a disinfection footprint. Then using this footprint, the ROI area is eroded avoiding collisions and finally, the rotating calipers path planner \cite{vasquez2020coverage} is modified to include a final tour that visits regions omitted.} To the authors' best knowledge, this is one of the first algorithms for automatically disinfecting urban areas that avoids collisions with the surrounding building. 

\textcolor{black}{
To validate the method, we have performed more than one hundred simulation experiments, testing our method with several ROI shapes and versus two related methods. The results show that the proposed method can plan successful paths despite the shape of the ROIs. In addition, the proposed method has the shortest computation time, making an on-board implementation feasible.
}



The rest of the article is presented as follows. \textcolor{black}{In section, \ref{sec:related work} we present the related work.} In section, \ref{sec:path_planning} we present our novel method for path planning. Section \ref{sec:experiments} describes the experiments that were carried out including a software in the loop simulation due to mobility restrictions. Finally, section \ref{sec:conclusions} presents our conclusion and future research directions.






\section{\textcolor{black}{Related work}}
\label{sec:related work}

Literature dealing with CPP for UAVs is as follows. \textcolor{black}{Meivel} \textit{et al.} \cite{me2016quadcopter} developed a system for spraying pesticides and fertilizers in open crop fields. The spraying UAV was controlled through manual flight plans. The UAV had a camera to capture remote sensing images, with the purpose of identifying green fields and the edges of the crop areas. The remote sensing images were analyzed by QGIS software to generate a map of the area. Also, Torres \textit{et al.} \cite{torres2016coverage} presented a method to solve the CPP problem to obtain a path that reduces battery consumption by decomposing non-convex polygons into convex areas. First of all, the UAV has a camera mounted. Then, it captures images from the ROI. Later, to perform the polygon coverage, the algorithm generates a back-and-forth movement or zigzag motion, the algorithm computes the distances between two straight movements using the overlap and the camera footprint. Finally, it returns a path that minimizes the number of UAV turns. Keller \textit{et al.} \cite{Keller_Coordinated} planned $G^2$ feasible paths for a UAV provided with a camera by concatenating $G^2$ curves in order to achieve persistent surveillance missions. An augmented A* algorithm was used to find a cycle sequence for the surveillance \textcolor{black}{ROI} This sequence was used in a B-spline curve generation algorithm to develop smooth paths that satisfy curvature constraints. For the same task, Xiao \textit{et al.} \cite{Xiao_Low} proposed a path planning algorithm based on the continuous updating of the virtual regional field and its local gradients. This virtual field incorporated a Boolean function which contains the information of the target regions and the obstacle information that forms a logical map. When a nonzero gradient at each point is in the regional field, the UAV finds the following target regions. Also, Stefas \textit{et al.} \cite{Stefas_vision} introduced an autonomous aerial system on a multi-rotor UAV navigating on an orchard for crop inspection. With a stereo camera yield data was collected from the \textcolor{black}{ROI.} For this, components for the UAV navigation, obstacle detection and avoidance based on vision, and CPP were developed. The autonomous aerial system used a global planner to enter, exit, and navigate to the next tree, also integrates a local planner to navigate on the tree rows. By their part, Freitas \textit{et al.} \cite{Freitas_capsule} combined area coverage planning with path planning for biological pest control by releasing capsules with a UAV. The locations to release the capsules in the infected areas were calculated using the hexagon as base geometry. The capsule was placed only if its center was within the infected areas. Then, a capsule was placed in the center of mass of each remaining sub-polygon of the previous step, whereas the path planning was achieved with ant colony optimization, guided local search, and Lin-Kernighan algorithms. Furthemore, Gao \textit{et al.} \cite{GAO2020} proposed a method to solve CPP for precise spraying in peach orchards. For that, a binocular color depth sensor was used to acquire video images. Then, a color depth fusion segmentation method based on the leaf wall area of the color depth images was proposed. Additionally, image erosion was used to delineate the two largest leaf wall areas as \textcolor{black}{an} ROI. The path of the spraying UAV was planned by detecting the midpoint of the ROI spacing as the end of the spray path.
Also, Dong \textit{et al.} \cite{Dong_tree} proposed an artificially weighted spanning tree coverage algorithm for the trajectory planning of flying robots. The robots simultaneously built their spanning tree, which grew toward the center of inertia of the uncovered area while stayed away from the trees of its partners. To go forward the tree, each robot iteratively evaluates the discovered cells and then selects the one having a maximum weight. According with this weighting, the selected cell is added to the spanning tree covering an area. By their part, Vasquez \textit{et al.} \cite{vasquez2020coverage} proposed a rotating calipers path planner that computes the full path to cover a convex ROI from a 2D perspective for disaster management or precision agriculture. The drone carried a pinhole camera during the mission and took photos of the ROI. The algorithm iteratively adds waypoints by intersecting a line with the ROI’s polygon. The algorithm requires parameters such as the polygon, an initial vertex, an adjacent vertex, an antipodal vertex, and the distance between flight lines. Furthermore, Skorobogatov \textit{et al.} \cite{Skorobogatov_Nonconvex} presented an algorithm open-source to divide any convex and non-convex polygon area into multiple parts, including any number of no-fly zones. Later, trajectories for the UAV were assigned. For the task, the UAV took pictures that were joined to obtain a complete map after a flight. The algorithm had as input a polygon defining the ROI, the initial positions of the UAV, the mission parameters, and a \textcolor{black}{convex} divisor function based on Hert and Lumelsky algorithm. Moreover, Tang \textit{et al.} \cite{Tang_R-DFS} introduced a CPP method based on a Region Optimal Decomposition (ROD) using a multi-rotor UAV in a maritime port for concave polygons. They applied the ROD to a Google Earth image of a port and combining the resulting sub-regions through an improved depth-first-search algorithm. Then, a genetic algorithm determined the traversal order of all sub-regions and connect the coverage paths. By \textcolor{black}{their} part, Yao \textit{et al.} \cite{Yao2016} proposed a mission assignment scheme for the farmland spraying problem by using multi-quadcopters. To solve the problem a mathematical model for the mission assignment and a sequential quadratic programming method to obtain the optimal solution were used. Thus, quadcopters can spray pesticides covering the farmland, but spraying is prohibited over areas covered by different quadcopters. Phung \textit{et al.} \cite{Phung_vision} formulated the inspection path planning problem as an extended Travelling Salesman Problem (TSP) for a UAV. They introduced a discrete particle swarm optimization (DPSO) algorithm using deterministic initialization, random mutation, and edge exchange to resolve the problem. The UAV has a CCD camera to detect potential defects or damages in the inspected area and, then, find the shortest path for the inspection of the planar surface. They used parallel computing for the velocity, position, and aptitude of the particles. The parallel program was implemented on a Jetson board mounted on the UAV using the MAVLink protocol. Also, Vasquez \textit{et al.} \cite{vasquez2018coverage} introduced a CPP method addressing disjoint areas based on the rural postman problem. The problem was resolved by optimizing the order of visits and computing the back-and-forth pattern. They converted the convex polygon into a reduced graph, and they treated the problem as the TSP, solving it by implementing a genetic algorithm. Finally, to compute the \textcolor{black}{back-and-forth} pattern, they found the visiting order, sorted the polygons and the set of vertices through their planner proposed.  Muliawan \textit{et al.} \cite{Muliawan2019} proposed a path planning approach for a UAV to carry out the spraying process in a plantation. A Modified Particle Swarm optimization (PSO) algorithm is used for the spraying task. The spraying process is carried out depending on the severity of disease of the plantation, which ranges from moderate to low.
Recently, Gonzalez \textit{et al.} \cite{Gonzalez_DE}  presented an approach to CPP for zigzag paths performed by a UAV in a three-dimensional environment. An optimization process based on the Differential Evolution (DE) algorithm is used in combination with the fast marching square scheduler. The UAV kept a fixed altitude to obtain images of a terrain, maintaining a homogeneous pixel size without overlapping. From the obtained images, a method to generate the zigzag path was used. Then, the DE algorithm optimizes zigzag path so that the steering angle of the UAV is optimal, ensuring a minimal distance cost. Campo \textit{et al.} \cite{Campo_crop} proposed a data acquisition system using a low-cost Lightweight UAV (LUAV) with a camera to cover areas of interest and obtain a continuous map in crops. The LUAV optimized the coverage paths using a heuristic strategy, where a waypoint for the navigation of the LUAV agent was the center of the footprint of the camera. Zuo \textit{et al.} \cite{Zuo_MILP} introduced a linear programming model to maximize coverage area and minimize coverage time for intelligence, surveillance, and reconnaissance missions with UAVs. In the first stage, a mission planner determined the search pattern, Point of Interest (POI), and the \textcolor{black}{ROI} for each UAV. In the second stage, the mission planner assigns some \textcolor{black}{ROIs} and flight paths for each UAV in the mission. Lastly, the aggregated mixed-integer linear programming for the path planning problem is solved by using the branch-and-bound algorithm in the CPLEX solver. Shang \textit{et al.} \cite{Shang_complex} presented a co-optimal CPP approach to generate an aerial inspection path that optimized coverage of a 3D surface and quality of the captured images, and reduced the computational complexity of the solver. The approach found the feasible paths for complete visual coverage. Later, a collection of the sampled paths were introduced in a PSO, which integrated the quality and efficiency of a coverage path in an objective function. Then, the calculated path was transferred to a flight trajectory using the rapidly exploring random tree to avoid obstacles. Tamayo \textit{et al.} \cite{Tamayo_cost} designed and implemented a software system to plan low-cost drone coverage for surveillance in agricultural or forested polygonal areas. The user specifies a start location and charging station (CS) locations. Then, the drone took a video of the area and creates a binary grid representation of the observed field. To minimize time to loading stations and field configurations, the branch-and-bound algorithm was implemented, which finds a minimal set of CSs that minimized distance and decomposed the field into Voronoi regions. After that, it computed a path using the modified TSP algorithm and constructs the paths. More recently, Biundini \textit{et al.} \cite{Biundini_framework} presented a framework for CPP for inspection with UAVs. They designed a metaheuristic algorithm based on point cloud data to inspect structures and coupling 3D reconstructions. First,  the camera of the UAV captured a moving image of the surface or ROI. The data of the structure is imported in point cloud or mesh format. Then, the data removed outliers and reduced the points of the optimization algorithm to identify the surface shape. After that, a genetic algorithm created a waypoint mission. Lastly, the path was sent to the UAV to start the flight, avoiding the UAV to waste energy. Popescu \textit{et al.} \cite{Popescu_flooded} designed a remote system to determine flooded areas through image processing with fixed-wing type UAVs. The system integrated terrestrial and aerial components. The \textcolor{black}{first} one is a coordinator at a distance of the UAV, which has a ground control station that communicated with more ground data terminals, via internet, through a network of nodes for data acquisition and communication. The second component corresponded to mobile nodes and the UAV, which must perform area coverage and image processing tasks during the mission. Thus, a deep neural network was used for texture analysis, color extraction, selection, and classification to provide the segmented image and the relative size of the flood. Liu \textit{et al.} \cite{Liu_energy} presented a navigation solution for UAVs working as a team. The UAVs were mobile base stations that flew around \textcolor{black}{an} ROI, performed the coverage task, and provided data services to a set of ground POIs. Each UAV needed to maintain connectivity with at least another UAV to avoid being isolated in the network. Then, a framework based on reinforcement learning was introduced to control each UAV. The UAVs were trained using environment state information, with the objective being to maximize the coverage for each vehicle, maximize the geographical fairness of all considered POIs or service points, and minimize total power consumption without leaving the edge of the area using a reward function. Godio \textit{et al.} \cite{Godio_bioinspired} introduced an approach based on bio-inspired neural networks to solve a CPP problem for surveillance and exploration in critical areas with a fleet of rotary-wing UAVs. The bio-inspired neural network was based on the propagation of the neuron dynamics of unvisited areas in all the map to guide the vehicles toward unexplored locations. Each neuron had a local cost and connection with neighboring neurons, considering the unvisited areas, obstacles, and UAV position in the fleet.

On the other hand, papers concerning to spraying UAVs without considering CPP are introduced in the following. Huang \textit{et al.} \cite{huang2009development} developed a low volume spraying UAV to apply crop protection products on specific growing areas indicated to the UAV through GPS coordinates or pre-programmed locations. Qin \textit{et al.} \cite{qin2016droplet} developed a similar system to spray pesticides in fields and mountainous areas. Suryawanshi \textit{et al.} \cite{suryawanshi2019design} introduced a system for agricultural fields which included a water pump connected to a fertilizer tank monitored with a sensor. When a certain threshold was reached the user was notified to fill the tank. In the work of Rao \textit{et al.} \cite{Rao2019} a semi-autonomous agricultural spraying system for pesticides was developed. The spraying system was connected to a control board, which in turn was connected to an Arduino board, which generates pulses to activate a DC motor and thus spray the pesticide. By they part, Uddin \textit{et al.} \cite{uddin2019unmanned} designed a system to clean windows of high-rise buildings with a UAV (Quadcopter) based on open source autopilot software and a LIDAR sensor. The drone sprayed water and a microfiber brush did the washing of the windows. Vempati \textit{et al.} \cite{vempati2019virtual} presented a UAV for painting 3D surfaces in desired locations with a spraying gun. A virtual reality interface is used to operate the system in a room, where the user could walk and interact with virtual objects. Zhang \textit{et al.} \cite{zhang2020effects} used a four-rotor drone to spray pesticides on sugarcane crop. The parameters took into account were spraying volume and height and speed of the flight. Lastly, some trade spraying UAVs are described in Table \ref{AspersoresC}.

\begin{table}[p]
	\caption{Main features of commercial sprinklers.}
	\begin{center}
		\begin{tabular}{ p{0.3\linewidth} p{0.6\linewidth}}
			\hline 
			Sprinkler & Features \\
			\hline
			Sprinkler SS600 Commercial Agriculture (Hexacopter) \cite{solutions}
			& 23 L tank. Covers up to 10 Ha/h. Integrates 4 nozzles. Flight time: 15-20min. 6 rotors of 100KV. Weight: 46.5kg. Payload: 24kg  \\ 
		
			Agras MG-1S Commercial agriculture (Octocopter)\cite{drone}  &  10 L tank. Covers up to 6 Ha/h. Integrates 4 nozzles. Flight time: 10-24min. 4 rotors of 130rpm. Weight: 8.0kg. Payload: 10kg \\ 
			
			DJI Agras T16 Commercial Agriculture (Hexacopter) \cite{tuxtla_2019} & 16 L tank. Covers up to 10 Ha/h. Integrates 8 nozzles. Flight time: 10-18min. 6 rotors of 75rpm. Weight: 18.5 kg. Payload:40.5kg. \\
			
			DRONEHEXA-AG Phytosanitary treatments (Hexacopter) \cite{user2020} & 16 L tank. Covers up to 2 Ha/v. Integrates 4-8 nozzles. Flight time: 10-18min. Weight: 12.4 kg. Payload:32kg\\
		    
			Sprinkler GAIA 160AG Agriculture (Hexacopter) \cite{foxtech2008} & 22.5 L tank. Covers up to 10 Ha/h. Integrates 4 nozzles. Flight time: 18min. 6 rotors of 100KV. Weight: 24 kg. Payload:46.5kg. \\ 
			\hline 	 
		\end{tabular} 
		\label{AspersoresC}
	\end{center}
\end{table} 

Lastly, other important contributions associated with flight controllers for UAVs and spraying methods are found in \cite{zhu2010development,xue2016develop,Spoorthi2017} and \cite{meng2019harvest,Potrino2018,meng2018effect,wang2019field}, respectively.

\section{Coverage path planning.}
\label{sec:path_planning}

\begin{figure}[tb]
    \centering
    \includegraphics[width=\linewidth]{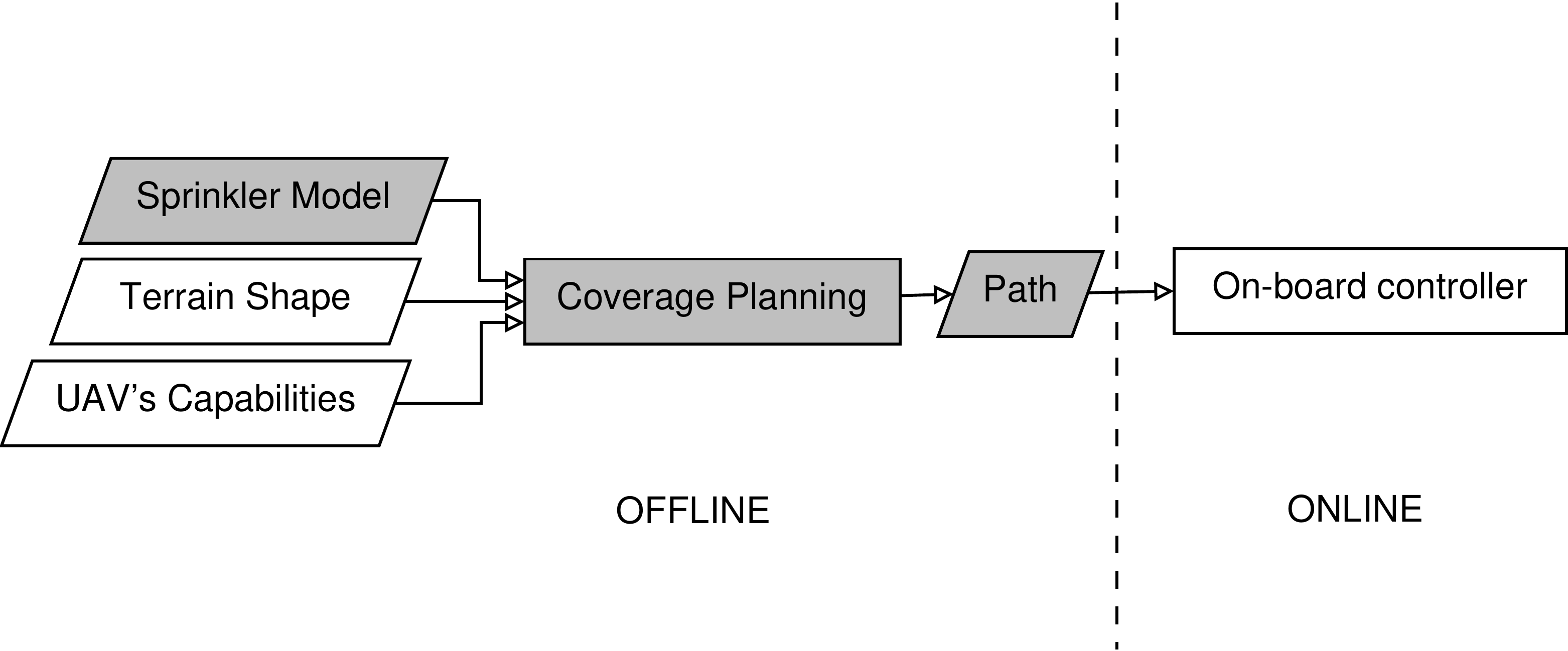}
    \caption{\textcolor{black}{General workflow diagram of coverage path planning for area disinfection. The required parameters are the sprinkler model, the terrain shape and the UAV specifications. Then, the planner uses these parameters to obtain a path. Finally, the computed path is used by the UAV to complete the mission. Our contributions are remarked in gray.}}
    \label{fig:plan}
\end{figure}

In this section, we present our method for automatically disinfect a two-dimensional area with an unmanned aerial vehicle that carries a sprinkler. The proposed method is composed of several parts, see Fig. \ref{fig:plan} where we show how such parts interact. First, we compute a sprinkler model using a regression approach that fits data obtained from the drop distribution to a paraboloid shape. Then, the sprinkler \textcolor{black}{model,} the ROI shape and the vehicle's capabilities are used to plan a path. The path, given by a set of waypoints, is loaded into the on-board controller that guides the vehicle through the set of waypoints. Below, we first introduce the sprinkler modeling, and then we present the coverage path planning.

\subsection{Sprinkler modeling}
\label{sec:modelado}

\begin{figure}[tb]
    \centering
    \includegraphics[width=\linewidth]{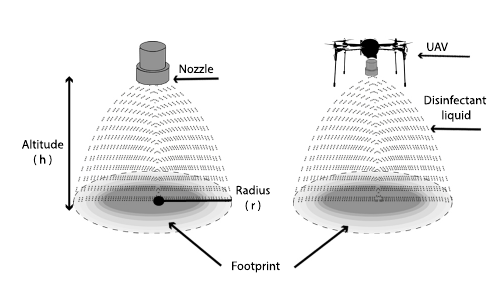}
    \caption{Paraboloid model of the disinfecting area. \textcolor{black}{Given a UAV that is flying at altitude $h$, the footprint of the sprinkler is given by a circle of radius equal to $r$.}}
    \label{fig:paraboloid}
\end{figure}

There are several ways to model a sprinkler system depending on the application and system characteristics \cite{grose1999mathematical}. In this work, we assume that only one sprinkler is mounted on the \textcolor{black}{UAV} and that the vehicle's altitude is relative \textcolor{black}{small.} For closed areas, the ceiling limits the \textcolor{black}{altitude; and for open areas the flight altitude should also be small,} otherwise the liquid effectiveness can be compromised. \textcolor{black}{See an illustration of the model in Fig. \ref{fig:paraboloid}.} Therefore, we model the kinematics of the sprinkler's coverage field as a rigid body governed by the equations of an inverse paraboloid:

\begin{equation}
    z = - Ax^2 - By^2 + h
    \label{eq:paraboloide}
\end{equation} \textcolor{black}{where $h$ is the drone altitude, $x$ and $y$ are coordinates on the plane, and the parameters $A$ and $B$ define the amplitude of the paraboloid.} Fig. \ref{fig:paraboloid} shows the proposed model. 

\begin{figure*}[tb]
		\centering
		\subfloat[Deterministic paraboloid.\label{subfig-1:parabolide}]{%
			\includegraphics[width=0.33\textwidth]{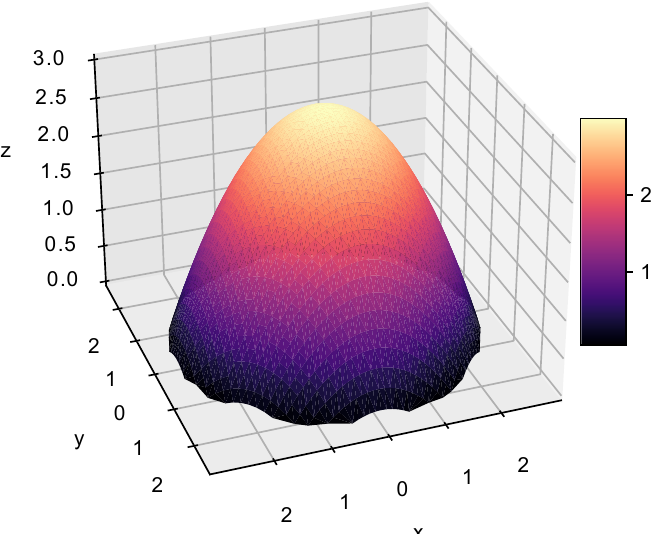}
		}
		\subfloat[Paraboloid with noise.\label{subfig-2:noise}]{%
			\includegraphics[width=0.33\textwidth]{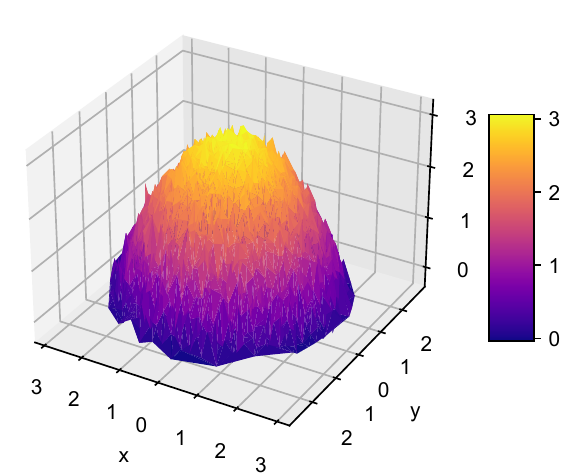}
		}
		\subfloat[Fitted paraboloid.\label{subfig-3:adjustment}]{%
			\includegraphics[width=0.33\textwidth]{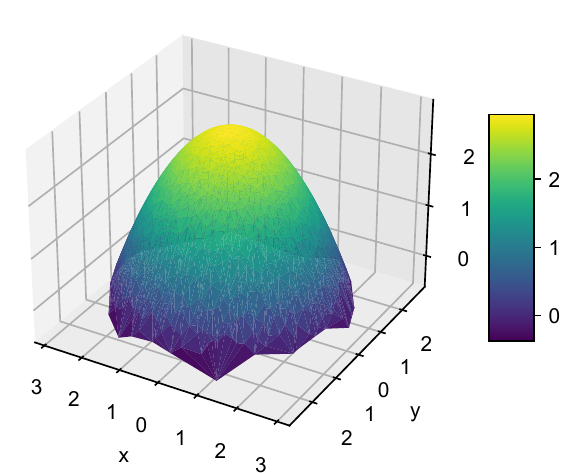}
		}
		\caption{Sprinkler model based on paraboloid equations for a UAV at three meters. The paraboloid bounds the volume covered by the sprinkler. Units are in meters. Figure best seen in color.}
		\label{fig:phases-paraboloids}
\end{figure*}

This model is closer to the reality w.r.t. previous models such as squares or cones. However, a deterministic model is still limited because in real sprinklers, the dispersion effect is much more complex given that many variables contribute to the drop's fall \cite{grose1999mathematical}. To insert such variability in our model, we consider the drop's fall as a stochastic process that affects the paraboloid shape, in consequence, we update \textcolor{black}{equation} (\ref{eq:paraboloide}) with a random noise.

\begin{equation}
    z' = - Ax^2 - By^2 + h + \epsilon_{\sigma}
    \label{eq:paraboloide2}
\end{equation} where $\epsilon_{\sigma}$ is a random error from a normal distribution with zero mean, $\mu = 0$, and standard deviation equal to $\sigma$. Fig. \ref{subfig-2:noise} shows an example of this noisy model. 

In many cases, the sprinkler is not characterized by the maker, however, a model fitting can be carried out so that the parameters $A$ and $B$ can be inferred using data from the drop's fall for a given sprinkler. In this method, we use the Levenberg-Marquardt Algorithm \cite{levenberg1944method} for finding the paraboloid parameters ($\hat{A}, \hat{B}$). Fig. \ref{subfig-3:adjustment} shows an example of a fitted model for a given set of observed drops. 


Next, let us assume that the terrain is horizontal, therefore, given the position of the vehicle, the spray footprint on the ground is determined by the intersection of the fitted paraboloid, equation (\ref{eq:paraboloide}), and the floor plane. In general, such footprint is described by an ellipsoidal shape. However, we restrict the intersection between the paraboloid and the floor to a circle. Look at Fig. \ref{fig:paraboloid}. The circle is the biggest one that fits into the ellipsoidal intersection. This restriction has several vantages: i) the plan is conservative, meaning that in the worst case more area is disinfected, ii) the path planning is simplified and iii) the processing efficiency is improved. A more detailed plan considering irregular footprints is left for future work. In summary, the footprint forms a circle, where its limit is given by the circumference with equation:

\begin{equation}
    r^2 = (x-a)^2 + (y-b)^2
\end{equation} \textcolor{black}{where $r$ is the circle's radius and (a, b) is the center of the circumference.} The proof of the intersection between the paraboloid and the plane is straightforward, so it is omitted. Then, from now on, the area that is sprayed inside the circle will be called as the drone's footprint. Once the vehicle is moving, the footprint evolves to a disinfected lane.

\subsection{Path planning}

Once the footprint has been stablished, the coverage path planning is defined as computing a vehicle's path in such a way that the ROI is completely covered by the footprint. The ROI is defined as a convex polygon, whose area is bigger than the footprint. Otherwise, the path is a single point. In addition, a constraint is that the vehicle must stay inside the ROI, e.g., for indoor scenes, the robot must avoid collision with doors and walls.

\begin{figure*}[tb]
		\centering
		\captionsetup[subfigure]{oneside,margin={0.5cm,0cm}}
		\subfloat[Original polygon (blue) and eroded polygon (red).\label{subfig-1:eroded}]{%
			\includegraphics[width=0.33\textwidth]{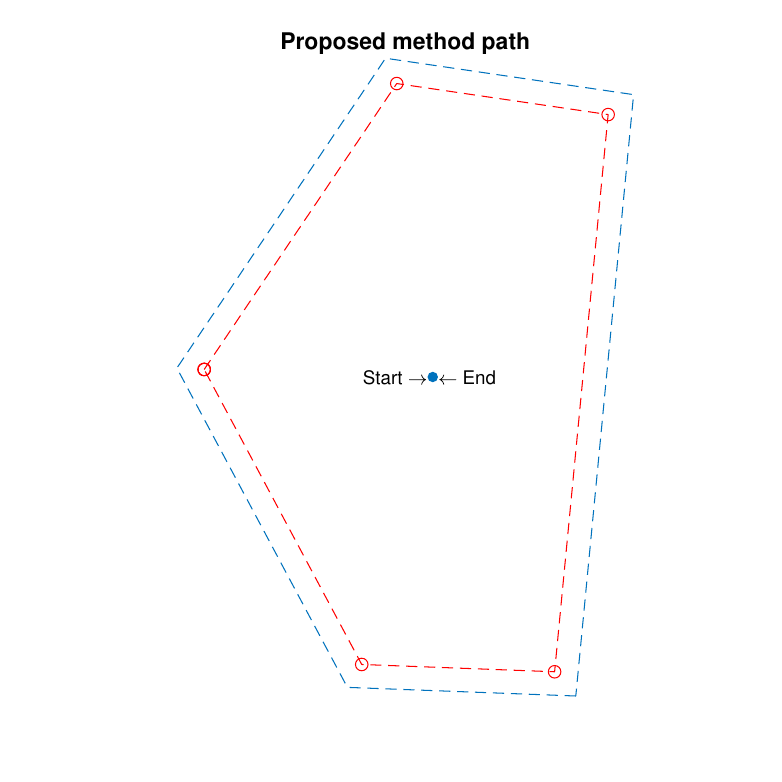}
		}
		\subfloat[First section of the coverage path forming a \textcolor{black}{back- and-forth} pattern.\label{subfig-2:path}]{%
			\includegraphics[width=0.33\textwidth]{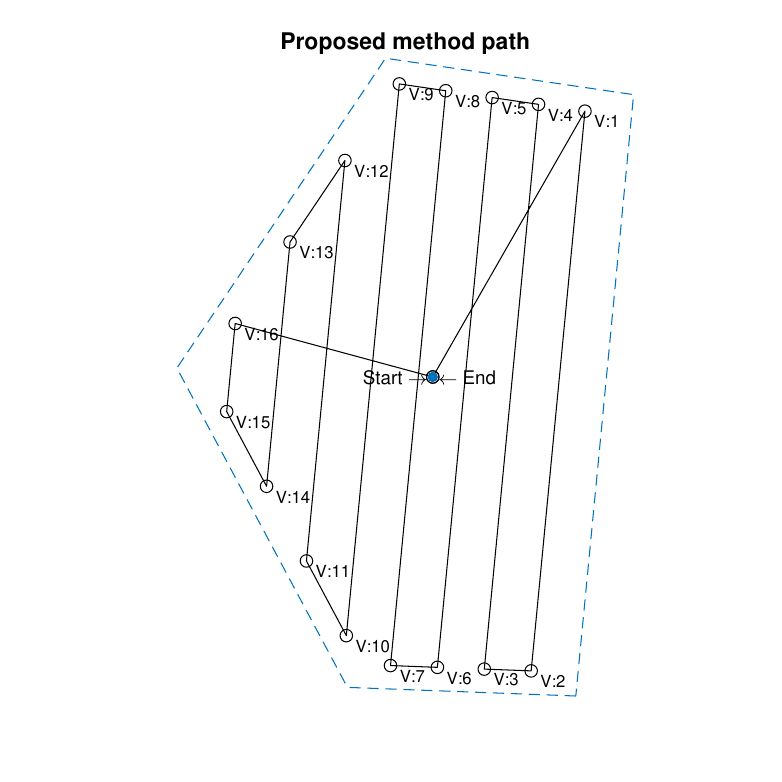}
		}
		\subfloat[Second section of the coverage path to fill missed areas.\label{subfig-3:fullpath}]{%
			\includegraphics[width=0.33\textwidth]{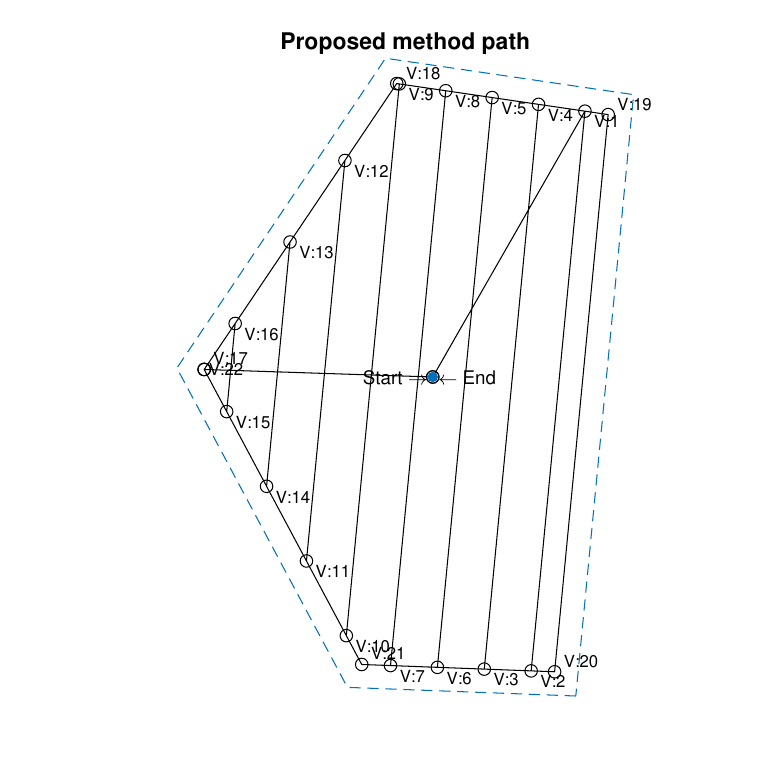}
		}
		\caption{Illustrations of the path computed by the proposed algorithm.}
		\label{fig:paraboloids}
	\end{figure*}

To solve the problem, we propose a method based on \textcolor{black}{back-and-forth} paths. This strategy simplifies the problem to finding a path composed by straight movements, called flight lines, in a similar way than a boustrophedon covers a terrain           \cite{choset1998coverage}. The method is summarized in algorithm \ref{alg:cpp} and described next. The requirements are the polygon ($M$), the footprint radius ($r$), the \textcolor{black}{takeoff point} ($s$) and the landing point ($e$). First, the method establishes the distance between flight lines, $\delta$, as two times the foot print radius. By setting $\delta=2r$, we guarantee that there is no gaps between flight of lines. Unlike previous literature       \cite{torres2016coverage}, we do not require overlap between flight lines since we assume that a single exposition to the spray is enough for disinfecting the area. Second, the ROI's polygon is eroded by a circular kernel of diameter equal to $r$. \textcolor{black}{Fig. \ref{subfig-1:eroded} shows an example of the original polygon and the result of the erosion.} The new polygon is called $M'$ \textcolor{black}{and it facilitates} the planning given that its frontiers can be reached by the UAV without collision. In addition, the erosion diameter is equal to $r$ because when the drone reaches the frontier of $M'$, the footprint is touching the frontier of $M$. Then, we use a modified version of the rotating calipers path planner (RCPP) \cite{vasquez2020coverage} for computing a path, $\rho$, that covers $M'$. \textcolor{black}{Fig. \ref{subfig-2:path} shows the path computed with the RCPP for the eroded polygon.} RCPP optimizes the flight of lines direction so that it gets the optimal path in linear time w.r.t. the number of vertices in the polygon. In our implementation, we connect flight lines over the frontier of $M'$, unlike the original method where the flight lines are connected by perpendicular sections, such perpendicularity might lead the UAV outside $M'$, in consequence the drone could collide with the environment. Finally, we modify the computed path by adding a tour of the $M'$ vertices before $e$, this tour covers some areas left in the path $\rho$ which is an implicit defect of the back-and-forth planners that deal with delimited regions \cite{galceran}. \textcolor{black}{Fig. \ref{subfig-3:fullpath} shows the inclusion of the final tour to the path.}

\begin{algorithm}[tb]
\SetAlgoLined
\KwData{Polygon ($M$), radius ($r$), starting point or home ($s$), final point($e$), distance between lines ($\delta$)}
\KwResult{Ruta ($\rho$)}
$\delta \leftarrow 2r$\;
$M' \leftarrow$  Erode($M$,$r$) \;
$\rho \leftarrow$ RCPP($M', \delta, s, e$)\;
$\rho \leftarrow \rho \cup$  Corners($M'$) \;
\caption{Coverage path planning for spraying drones.}
\label{alg:cpp}
\end{algorithm}

\section{Experiments}
\label{sec:experiments}

\textcolor{black}{
In this section, we present several experiments for validating our method. First, we test the method with several polygon shapes, where each shape is defined by a fixed number of vertices but the size of the edges might vary. In that experiment, we compare our method versus two state-of-the-art approaches. Second, we simulate the execution of the proposed method in a realistic scenario, this simulation was set as close as possible to a real disinfection. So far, validating our method with real world experiments has been impossible due to current travel and mobility restrictions imposed by COVID19.
}

\textcolor{black}{
Our implementation was done using several programming frameworks. The planners were implemented in MatLab; they read the target polygon and ouput the path. A second implementation in C++ reads the path and communicates with the the Gazebo simulator to perform a software in the loop real time simulation. 
}

\subsection{Random polygons experiment}


\textcolor{black}{
In this experiment, the proposed method is tested and compared versus two state-of-the-art methods. The goal is to show that the method can deal with the polygons despite their shape. The first of the compared methods is the proposed by Choset el at. \cite{choset1998coverage}, where vertical back-and-forth paths are computed to cover the regions. The second compared method is the one proposed by Torres et al. \cite{torres2016coverage}; in that method, the flight lines are oriented perpendicularly to the minimum polygon width. For the comparison, we have performed a statistical analysis where several polygons of different shapes are tested with each method. In detail, we randomly create five polygons for each shape from a set of eight shapes. The set of shapes includes polygons from three vertices to ten vertices. In consequence, 120 tests were carried out in real-time simulations.
}

\begin{figure}[tb]
    \centering
    \subfloat[Tree vertices \label{3_Proposed_method}]{%
			\includegraphics[width=0.20\linewidth]{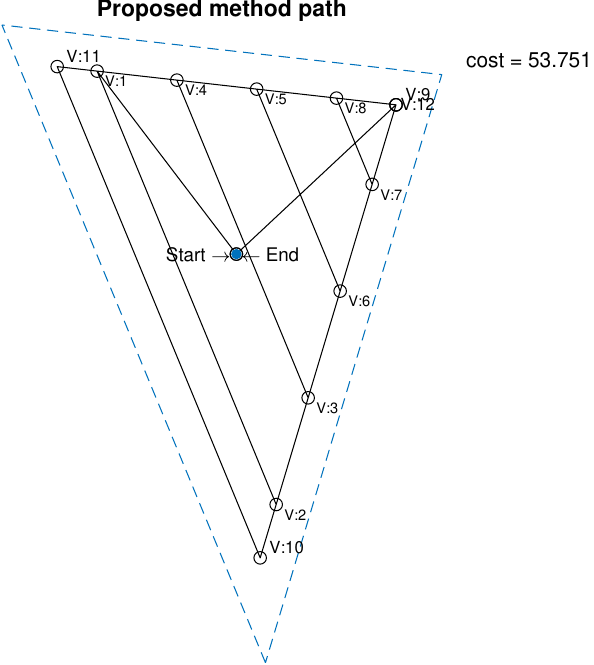}
		}
		\subfloat[Four vertices \label{4_Proposed_method}]{%
			\includegraphics[width=0.20\textwidth]{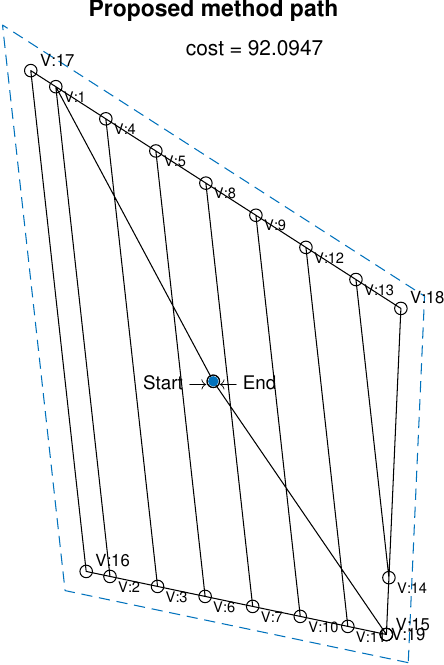}
		}
				\subfloat[Five vertices \label{5_Proposed_method}]{%
			\includegraphics[width=0.20\textwidth]{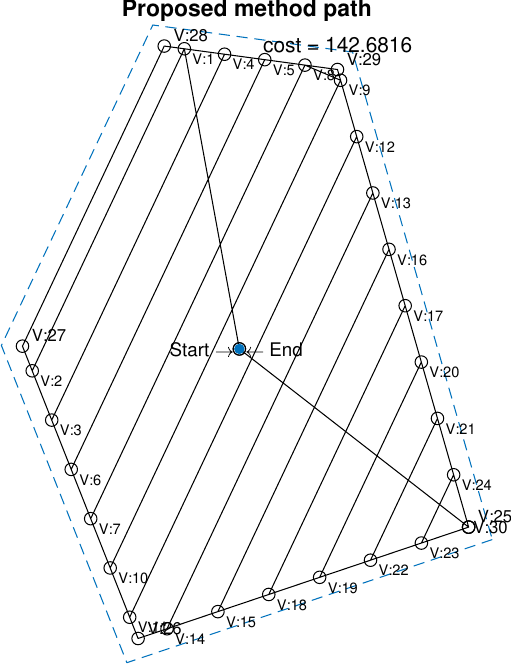}
		}
		\subfloat[Six vertices \label{6_Proposed_method}]{%
			\includegraphics[width=0.20\textwidth]{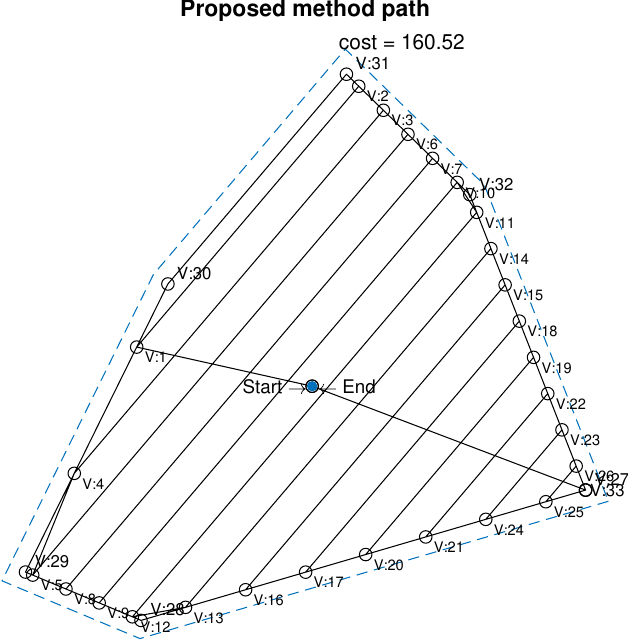}
		}

			\subfloat[Seven vertices \label{7_Proposed_method}]{%
			\includegraphics[width=0.20\linewidth]{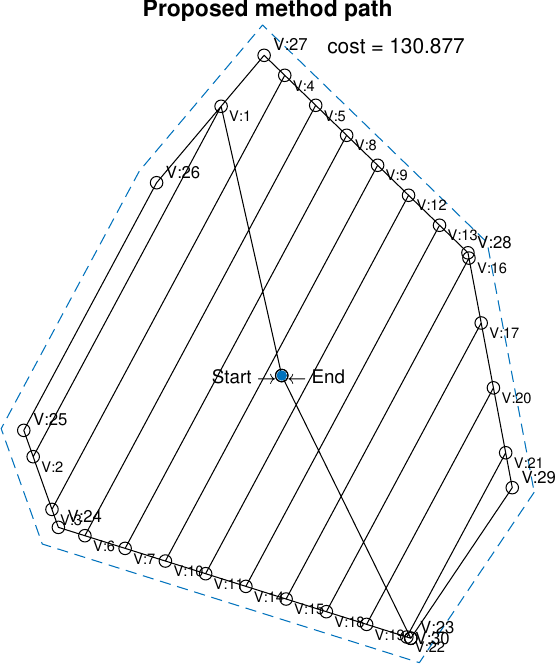}
		}
		\subfloat[Eight vertices \label{8_Proposed_method}]{%
			\includegraphics[width=0.20\linewidth]{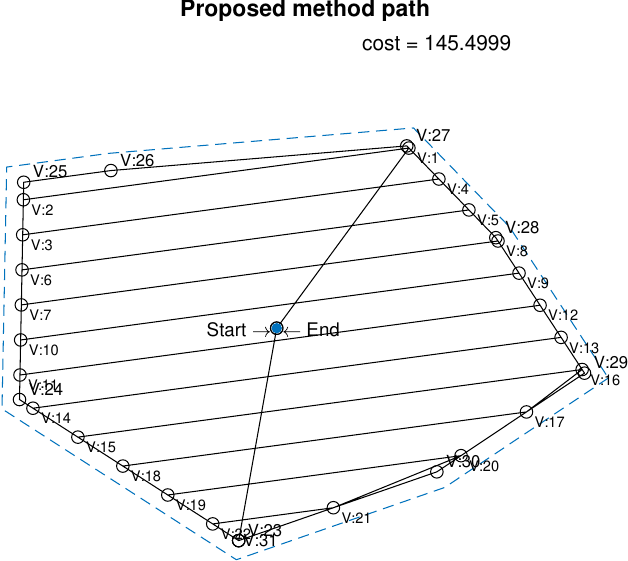}
		}
			\subfloat[Nine vertices \label{9_Proposed_method}]{%
			\includegraphics[width=0.20\linewidth]{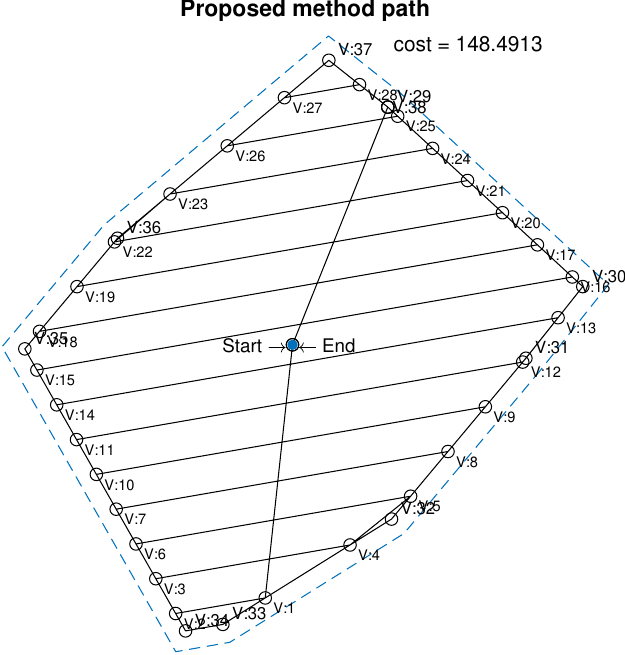}
		}
		\subfloat[Ten vertices \label{10_Proposed_method}]{%
			\includegraphics[width=0.20\linewidth]{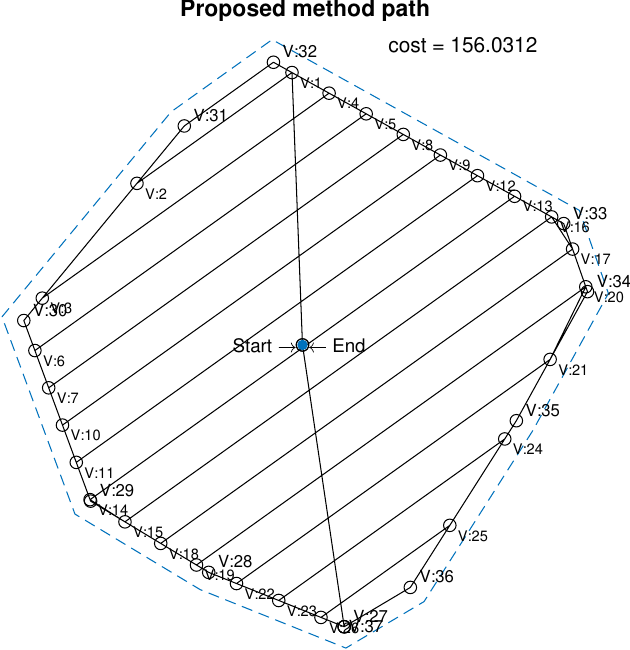}
		}
    \caption{\textcolor{black}{Examples of the polygons that were tested and their corresponding paths with the proposed method. The polygons are drawn in dotted blue lines and the paths are drawn in solid black lines. Each polygon is one of five random polygons that where created for a given shape. The shapes include polygons from three vertices to ten vertices.}}
    \label{fig:poly_examples}
\end{figure}



\textcolor{black}{
The polygons used in the experiments have random shapes but the number of vertices is set by the user. Their shapes are random variations of regular polygons. Let us define a regular $n$-sided polygon as a polygon with $n$ vertices and $n$ sides that is equiangular and equilateral; in addition, the radius is the distance between each vertex and the center. The variation consists of drawing the vertices of the polygon counter-clockwise but adding a random amount to the angle w.r.t. the previous vertex (this eliminates the equiangular property) and to the radius. The random variables are drawn from a uniform distribution between zero and a maximum amount. In particular, these experiments have a target radius of 20 meters with a maximum radius variation of 1 meter and a maximum angle variation of 1 radian. In Fig. \ref{fig:poly_examples}, we display some examples of the polygons that were tested.
}

\textcolor{black}{
Table \ref{parameterspolygon} summarizes the average results of the experiments. The columns in the table display the number of vertices of the polygons tested, and the path distance (measured from the computed path) and the simulated flight time (measured by Gazebo simulator) of the compared methods.
}

\begin{figure}[tb]
		\centering
		\subfloat[Computed path. \label{subfig-3:fullpath3}]{%
			\includegraphics[width=0.30\linewidth]{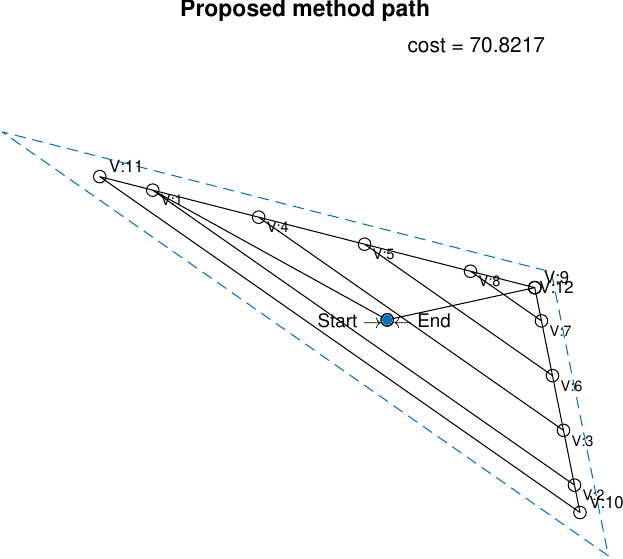}
		}
		\subfloat[Torres' path.  \label{path_torres3}]{%
			\includegraphics[width=0.30\textwidth]{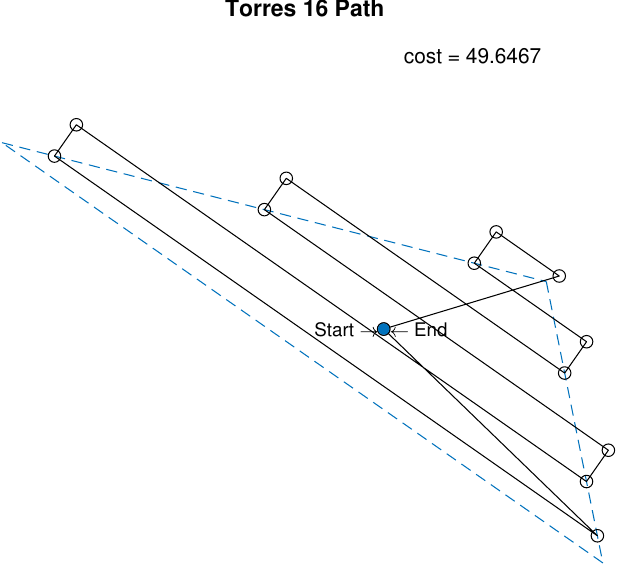}
		}
				\subfloat[Choset' path. \label{subfig-3:fullpath4}]{%
			\includegraphics[width=0.30\textwidth]{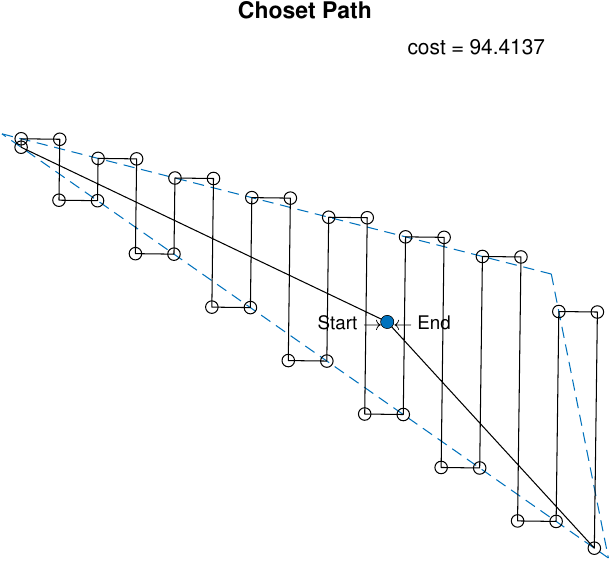}
		}
		
			\subfloat[Computed path. \label{subfig-3:fullpath6}]{%
			\includegraphics[width=0.30\linewidth,height=4.4cm]{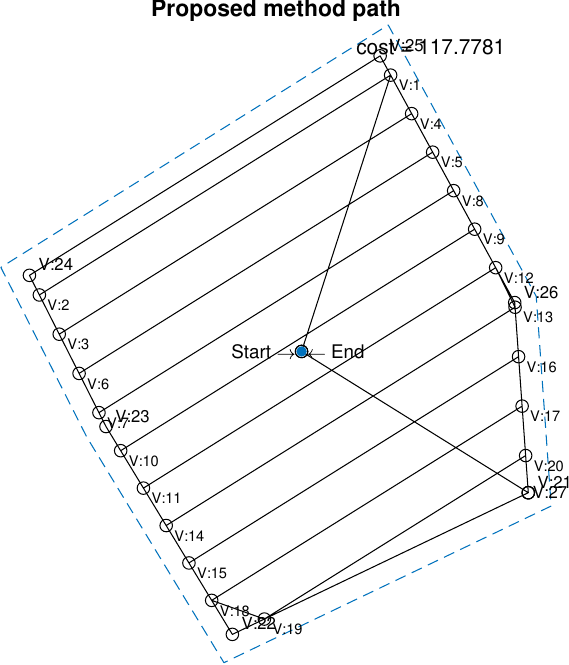}
		}
		\subfloat[Torres' path.   \label{path_torres6}]{%
			\includegraphics[width=0.30\linewidth]{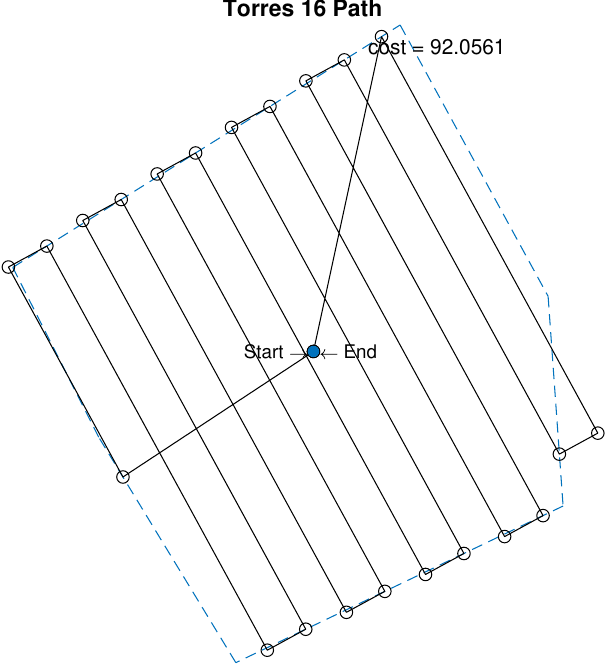}
		}
			\subfloat[Choset' path. \label{subfig-3:fullpath8}]{%
			\includegraphics[width=0.30\linewidth]{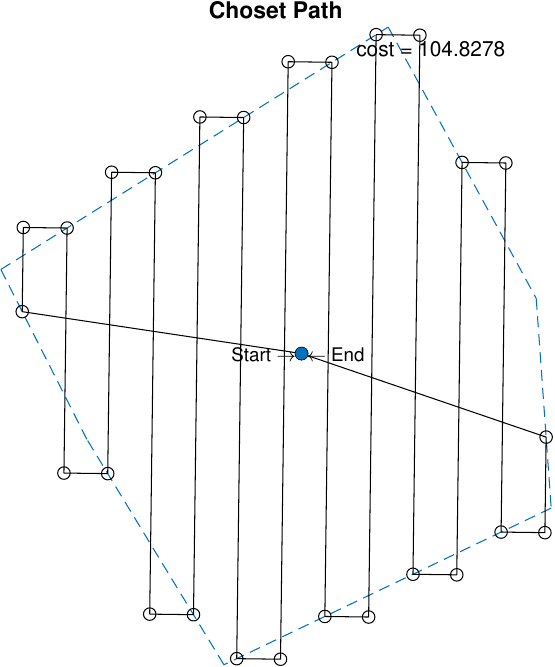}
		}
		


		\caption{Comparison of computed paths by our proposed approach versus Torres et al. \cite{torres2016coverage} and Choset et al. \cite{choset1998coverage}. In all cases, the target polygons are draw with dashed lines and the path is drawn with solid lined. Figure best seen in color.}
		\label{fig:pol3}
	\end{figure}

\textcolor{black}{
From the experiments, we can observe that the algorithm works for different shapes and different number of vertices. In all cases, the paths generated by our method are inside the target polygon avoiding collisions. This behaviour can be observed in Fig. \ref{fig:pol3} where we compare the paths for three random polygons. For the other two methods, the paths are outside the target polygon, so the vehicle can collide or it can get into restricted areas. 
}

\textcolor{black}{
The processing time for the proposed approach is $0.06$ seconds in average, while the time for Torres's algorithm is $0.136$ seconds and the time for the Choset's algorithm is $0.60$ seconds on average. The reason why our method takes less time to compute the path is due to the rotating calipers path planner, which computes the path with linear complexity with respect to the number of vertices. This fact has already been proved in \cite{vasquez2020coverage}.
}

\textcolor{black}{
A natural consequence of our method is that the path length is increased given that it performs a final tour around the target polygon for covering the missed regions. In Table \ref{parameterspolygon}, we can observe this increment in distance, which is small if it is compared versus the Choset's method.
}

\begin{table}[htb]
   	\renewcommand{\arraystretch}{1.3}
	\caption{Mean results for the tested polygons. The columns indicate: $N$ for number of vertices, ``D total" for total traveled distance and ``Sim. $t$" for simulation flight time. Distances are expresses in meters and time is expressed in minutes.}
	\begin{center}
	\begin{tabular}{r r r r r r r}
	\hline
	Vertices
	&
	\multicolumn{2}{c}{Proposed method}
    &
	\multicolumn{2}{c}{ Torres\cite{torres2016coverage}}
	&
	\multicolumn{2}{c}{ Choset\cite{choset1998coverage}} \\
	\hline
			 $N$ &$D$ total & Sim. $t$ &$D$ total & Sim $t$ &$D$ total & Sim. $t$ \\ 
			\hline 
		     3&76.18 &04:09 &56.33 &03:47&85.85 &04:16\\ 
				 
			 4&98.18 &05:18  &  75.50m &04:48 &96.55 &  05:13\\
			
		     5&126.32 & 06:47&100.62  &06:10 &110.26& 06:26\\
		    
		    6 & 128.54 &07:01& 103.43 &06:29 &114.91 & 06:46 \\
		    
		    7 &149.65 & 07:59& 120.33 &07:21 &131.02 &07:30\\ 
		    
		    8 &153.14 &08:36  &128.66 &07:58 & 143.22 & 08:07\\
		    
		    9 & 143.90  & 07:45 &116.68 &07:10 &130.10 & 07:35
 \\
		   10 & 155.09 &08:40 & 128.96 &07:58&142.65 & 08:16 \\
		   \hline 
		\end{tabular} 
		\label{parameterspolygon}
	\end{center}
\end{table} 

\subsection{3D simulation}

\begin{figure}[tb]
		\centering
		\subfloat[Planned path.\label{subfig-5:sim1-1}]{%
			\includegraphics[height=0.43\textwidth]{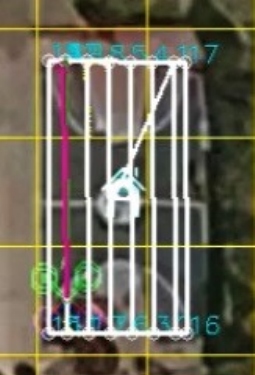}
		}
			\subfloat[Online excecution of the path.\label{subfig-5:sim1-2}]{%
			\includegraphics[height=0.43\textwidth]{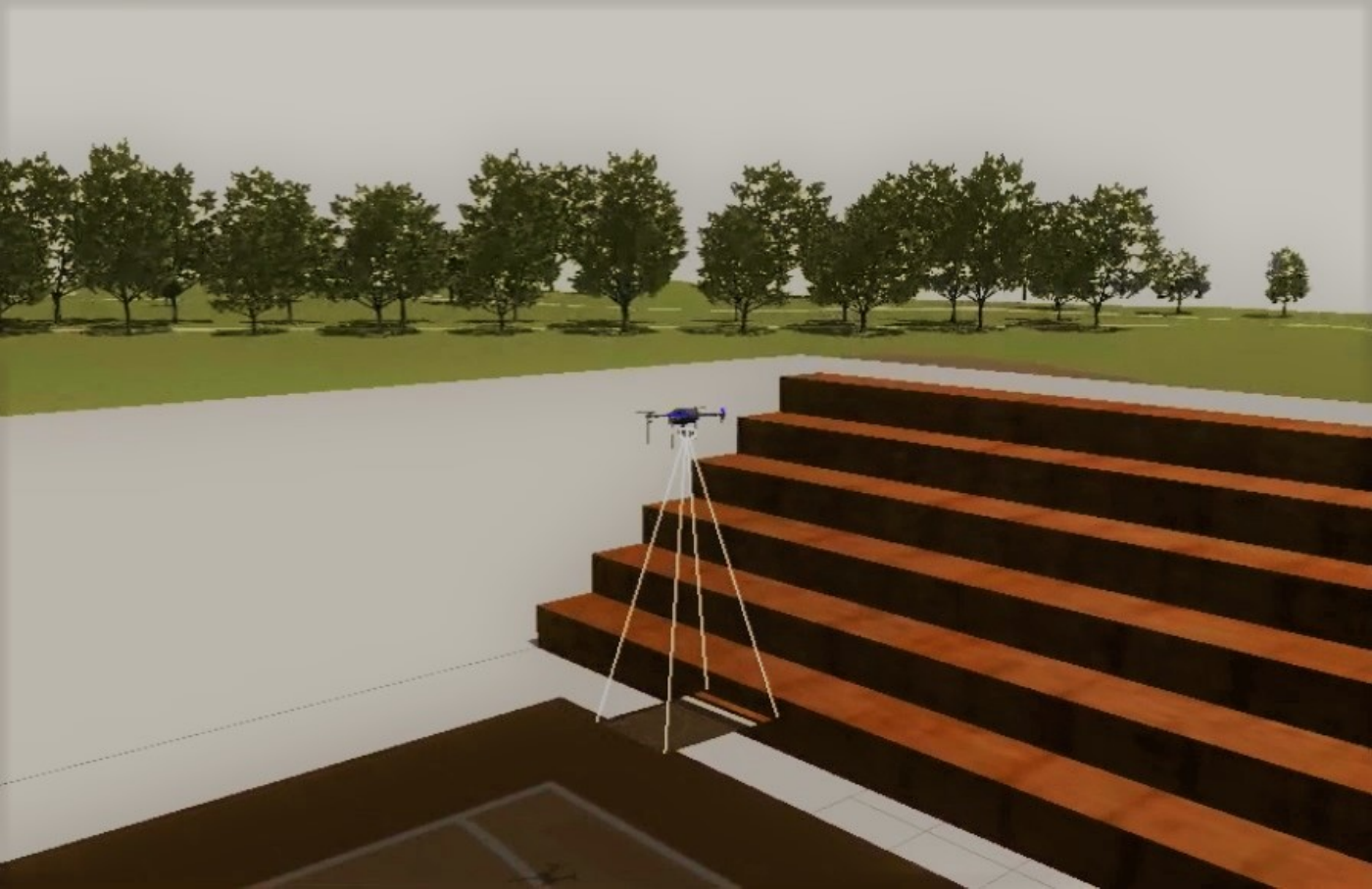}
		}
		\caption{Software in the loop simulation (SITL) where the vehicle in Gazebo simulator is commanded through the MAVLink protocol. Once the path is loaded in the on-board controller, it flights autonomously.}
		\label{fig:sims}
\end{figure}

The objective of this experiment is to validate that the method works in a more realistic simulation. See Fig. \ref{fig:sims}. Due to the current situation, it was not possible to run the experiments in a real vehicle. However, the Gazebo simulator was configured with closest to reality parameters. In addition, we want to provide a qualitative display of the planned task.

We draw a polygon on the edge of the area surrounding one of the Bicentennial Park basketball courts located in Mexico City using the Mission Planner software. Later, the coordinates define the ROI. Next, the coordinates are converted from geodetic coordinates (latitude, longitude, and height) to NED coordinates (North, East, and Down). The erosion function in the algorithm is in charge of reducing the border of the area that will depend on the radius of the sensor, and once it has the area, it continues to the creation of the flight lines. However, it is required to visit all the areas so that as much as possible is covered. The algorithm adds the nodes of all the polygon vertices to cover areas without visiting. Then a conversion from NED coordinates to geodetic coordinates is performed again to add these coordinates to a waypoints file. The Gazebo simulator uses the waypoints file to carry out the path over the court.

Subsequently, for this simulation, SITL was used in collaboration with the Gazebo 9 simulator to see the area in 3D. In this simulation, the Quadcopter 3DR Iris was used. The height set during the simulation was 10 meters. The speed of the drone is \textcolor{black}{2m/s.} \textcolor{black}{The starting point is set} with the coordinates 19.467468, -99.193345. Additionally, a basketball court model made for Gazebo allows to evaluate the behavior of the UAV in a virtual environment.

It was observed in the experiment that the vehicle covered most of the basketball court area without going out of the contour, thus avoiding some collision. In total, it covered an area of 59.4966 meters in a time of 3:46 minutes. \textcolor{black}{The reader can see the execution of this experiment in the following video: \url{https://youtu.be/2ERkEagnCEM}.}

\subsection{Discussion}


\textcolor{black}{
The proposed method has three advantages over the compared methods: i) avoiding potential collisions since the paths are inside the target areas, ii) providing more realistic modeling of the sprinkler and iii) shorter computation times.} Avoiding potential collisions is one of the main features of the planner, since its application is oriented to low altitude missions. The processing time of the proposed approach is shorter because it is based on the rotating calipers algorithm to compute the inner path. With respect to traveling distance, our method computes a larger path since it performs a final tour around the polygon to cover missed areas.

\section{Conclusions}
\label{sec:conclusions}

A method for planning two-dimensional area disinfection with an \textcolor{black}{UAV} has been presented. Our method extends previous research by including a more precise sprinkler modeling and restricts the vehicle flight inside the \textcolor{black}{Region of Interest.} In consequence, it improves coverage and avoids potential collisions. Through several experiments, we have validated our method; in those experiments, we observed that the vehicle maximizes the area of disinfection despite the area shape. The method has shown to be efficient enough for being implemented in on-board systems with low computational capacity. For future work, we plan to extend our method for non-convex polygons and to carry out real-world experiments.


\bibliographystyle{elsarticle-num} 
\bibliography{bibliografia}





\end{document}